\documentclass[letterpaper, 10 pt, conference]{ieeeconf}  % Comment this line out if you need a4paper
\usepackage{apacite}

\usepackage{float}
\usepackage[colorlinks=true,linkcolor=black,anchorcolor=black,citecolor=black,filecolor=black,menucolor=black,runcolor=black,urlcolor=black]{hyperref}

\IEEEoverridecommandlockouts                              % This command is only needed if 
\usepackage{algorithm2e}
\usepackage{graphicx}
\usepackage[skip=4pt]{caption}
\usepackage{amsmath}
\usepackage{amssymb}
\usepackage{graphicx}
\usepackage{float}
\usepackage{subfigure}
\usepackage{verbatim}
\makeatletter
\let\NAT@parse\undefined
\makeatother
\usepackage[comma]{natbib}  %
\setcitestyle{aysep={}}
\setcitestyle{authoryear,open={(},close={)},aysep={}}

\newcommand{\Aspace}{\mathcal{A}}
\newcommand{\Sspace}{\mathcal{S}}

\newcommand{\MDP}{\mathcal{M}}
\newcommand{\E}[2]{\operatorname{\mathbb{E}}_{#1}\left[#2\right]}
\newcommand{\R}{\mathbb{R}}
\newcommand{\state}{\mathbf{s}}
\newcommand{\st}{\state_t}
\newcommand{\stp}{\state_{t+1}}

\newcommand{\action}{\mathbf{a}}
\newcommand{\at}{\action_t}
\newcommand{\atp}{\action_{t+1}}

\newcommand{\D}{\mathcal{D}}

\title{\LARGE \bf
Asynchronous Reinforcement Learning\\ for Real-Time Control of Physical Robots}

\author{Yufeng Yuan$^{1}$ \& A. Rupam Mahmood$^{1,2}$% <-this % stops a space
\thanks{$^{1}$Department of Computing Science, University of Alberta, Edmonton, Canada, T6G 2E8.
        {\tt\small \{yufeng10, armahmood\}@ualberta.ca}  $^{2}$CIFAR AI Chair, Alberta Machine Intelligence Institute (Amii)}
}

\begin{document}

\maketitle
\thispagestyle{empty}
\pagestyle{empty}

%%%%%%%%%%%%%%%%%%%%%%%%%%%%%%%%%%%%%%%%%%%%%%%%%%%%%%%%%%%%%%%%%%%%%%%%%%%%%%%%
\begin{abstract}

An oft-ignored challenge of real-world reinforcement learning is that the real world does not pause when agents make learning updates. As standard simulated environments do not address this real-time aspect of learning, most available implementations of RL algorithms process environment interactions and learning updates sequentially. As a consequence, when such implementations are deployed in the real world, they may make decisions based on significantly delayed observations and not act responsively.
Asynchronous learning has been proposed to solve this issue, but no systematic comparison between sequential and asynchronous reinforcement learning was conducted using real-world environments.
In this work, we set up two vision-based tasks with a robotic arm, implement an asynchronous learning system that extends a previous architecture, and compare sequential and asynchronous reinforcement learning across different action cycle times, sensory data dimensions, and mini-batch sizes.
Our experiments show that when the time cost of learning updates increases, the action cycle time in sequential implementation could grow excessively long, while the asynchronous implementation can always maintain an appropriate action cycle time.
Consequently, when learning updates are expensive, the performance of sequential learning diminishes and is outperformed by asynchronous learning by a substantial margin.
Our system learns in real-time to reach and track visual targets from pixels within two hours of experience and does so directly using real robots, learning completely from scratch. Our code is available at: \\ \url{https://github.com/YufengYuan/ur5_async_rl}.

\end{abstract}

%%%%%%%%%%%%%%%%%%%%%%%%%%%%%%%%%%%%%%%%%%%%%%%%%%%%%%%%%%%%%%%%%%%%%%%%%%%%%%%%
\section{INTRODUCTION}

Deep reinforcement learning has been a promising approach to online learning of robotic control~(e.g.,~\citealt{schulman2015trust}, \citealt{duan2016benchmarking}, \citealt{schulman2017proximal}, \citealt{haarnoja2018soft}). 
However, except for a few recent works~(e.g.,  \citealt{kalashnikov2018qt},
\citealt{mahmood2018benchmarking}, \citealt{schwab2019simultaneously}), most works focus on training offline or in simulations instead of real-time learning with physical robots.
This research gap is primarily attributed to the properties of the real world, such as the complexity, partial-observability, and slow data collection~\citep{dulac2020empirical}. 

An additional challenge when applying reinforcement learning in the real world is the sequential computation of environment interactions and learning updates in most available implementations.
Such computation arrangement is appropriate in simulated environments that are internally paused while agents make learning updates.
However, in real-world environments, such sequential computation could delay the observation and the execution of the next action when learning updates are in progress, potentially prolonging the effective length of the time step or \emph{action cycle time} and reducing the responsiveness of the agent. 
In practice, this issue can be considerably exacerbated when high-dimensional data, such as images, are combined with algorithms with expensive per-step updates such as Soft Actor-Critic~\mbox{(SAC, \citealt{haarnoja2018soft})} or Deep Deterministic Policy Gradient~\mbox{(DDPG,  \citealt{lillicrap2015continuous})}. 
Although prior learning architectures~\citep{haarnoja2018learning}
addressed this issue by running learning updates and environment interactions asynchronously, no systematic study of the benefits of asynchronous learning over sequential learning has been conducted on real-world robotic tasks.

In this work, we study the advantages of asynchronous learning over sequential learning in real-world reinforcement learning under four specific scopes: 1) low-level control policy learning, 2) learning in real-time, 3) learning from scratch, and 4) learning from images. We implement two vision-based tasks with a UR5 robotic arm using the SenseAct framework~\citep{mahmood2018benchmarking}.
The first task with a stationary target can be solved by a coarse control policy, while the second task with a constantly moving target requires a finer control policy. 
The existing asynchronous learning architecture, based on Soft Actor-Critic~\citep{haarnoja2018soft}, can be sufficient for low-dimensional data but may not be efficient enough when learning from high-resolution images. Thus, we extend this architecture by separating replay buffer sampling and gradient updates inside the learning update process into two parallel processes.
Such parallel systems, together with multiple actors interacting with multiple instances of the environment, have been proposed and used in large-scale simulations~(e.g., \citealt{nair2015massively}, \citealt{espeholt2018impala},  \citealt{barth2018distributed}), but the benefit of these architectural variations has not been analyzed previously on real-world robots.

\iffalse
In this work, we implement two vision-based tasks with a physical robotic arm to 
compare sequential and asynchronous learning with high-dimensional observations and expensive per-step updates in the real world.
The two tasks are based on the SenseAct framework~(\citealt{mahmood2018setting}), consisting of a UR5 robotic arm performing reaching and tracking behaviors, a monitor displaying the target, and a wrist-mounted RGB camera capturing images. The first task with a stationary target can be solved by a coarse control policy, while the second task with a constantly moving target requires a finer control policy. Our asynchronous learning architecture, based on Soft-Actor-Critic~(\citealt{haarnoja2018soft}), extends the existing architecture~(\citealt{haarnoja2018learning}) that runs environment interactions and learning updates in parallel. 
Such architectures can be sufficient for low-dimensional data but may not be efficient enough when learning from high-resolution images. Thus, we extend this architecture by separating replay buffer sampling and gradient updates inside the learning update process into two parallel processes.
Such parallel systems, together with multiple actors interacting with multiple instances of the environment, have been proposed and used in large-scale simulations~(e.g., \citealt{nair2015massively}, \citealt{espeholt2018impala},  \citealt{barth2018distributed}).
However, the benefit of these architectural variations has not been analyzed previously on real-world robots.
\fi

We conduct systematic experiments to compare sequential learning and the two variants of asynchronous learning systems across different action cycle times, sensory data dimensions, and mini-batch sizes. 
Our experimental results show that sequential and asynchronous learning can perform similarly when learning updates are relatively cheap.
However, when the learning updates become expensive due to higher resolution of images or larger mini-batch size, the performance of sequential learning diminishes as its action cycle time is prolonged excessively, which significantly reduces the agent's responsiveness and the number of learning updates.
Consequentially, the asynchronous learning system performs substantially better than sequential learning in this case.
We also show that when the mini-batches are considerably large, the additional parallelization of replay buffer sampling and gradient updates provides significant performance improvement. 
Besides the better performance, two clear advantages of asynchronous learning are illustrated through our experiments.
First, the minimal action cycle time achievable by asynchronous learning can be much shorter than that of sequential learning, which provides more flexibility when learning in the real world. Second, unlike sequential learning,  a constant action cycle time can always be maintained by asynchronous learning regardless of the time cost of learning updates.
Our real-world robotic system learns effective reaching and tracking behaviors from pixels starting completely from scratch. Our publicly available implementations of the tasks and the learning system may enable reproducibility and accelerate further advancement of real-world and real-time robot learning from images.

\section{RELATED WORKS}\label{hypothesis}
\subsection{Reinforcement Learning from Pixels}
Reinforcement learning from pixels~\citep{mnih2013playing} usually needs a prohibitively large amount of data due to the relatively sparse reward signal, high-dimensional data, and partial-observability. To alleviate this issue, one common approach is to add auxiliary tasks, such as self-supervised prediction~\citep{jaderberg2016reinforcement}, auto-encoder reconstruction~\citep{yarats2019improving}, and contrastive learning~\citep{stooke2020decoupling}, to provide more training signals to visual representation. A complimentary but highly effective approach is to add random augmentations to image input, such as random cropping proposed by~\citet{kostrikov2020image} and \citet{laskin2020reinforcement}, to regularize representations by learning an augmentation-invariant critic. However, most of the works consider simulated environments with a stationary overhead camera.
In contrast, our setting has the camera mounted at the end of a physical robotic arm.

\subsection{Reinforcement Learning with physical Robots}
Reinforcement learning has been applied to robotic tasks, such as motor skills~\citep{lange2012autonomous}, grasping objects~\citep{kalashnikov2018qt}, in-hand object manipulation~\citep{andrychowicz2020learning}, door opening~\citep{gu2017deep}, and locomotion~\citep{haarnoja2018learning}. To successfully solve these tasks, different methods have been proposed to overcome the real-world challenges, such as ensuring critical constraints are never violated~\citep{dalal2018safe}, removing manual reset by learning a resetting policy~\citep{eysenbach2017leave}, removing hand-engineered reward functions~\citep{zhu2020ingredients}, utilizing prior knowledge in simulations~\citep{peng2018sim}, and using auxiliary tasks~\citep{schwab2019simultaneously}.
Although some of these works already utilize asynchronous learning or distributed learning, a careful study of the difference between sequential and asynchronous learning is still missing, and few of these works make their asynchronous implementation publicly available.

\section{Background}\label{background}
\subsection{Markov Decision Process} 
We formulate our continuous control task as an finite-horizon Markov decision process~(MDP, \citealt{bellman1958dynamic}) described as a tuple $\MDP = (\Sspace, \Aspace, p, r, \gamma, d_0)$. In this tuple, $\Sspace$ is the set of all states and $\Aspace$ is the set of all actions, which are both continuous in our case. Moreover, $p = Pr(\stp |\st, \at)$ is the transition dynamics, which captures the probability distribution over the next state $\stp \in \Sspace$ given the current state $\st \in \Sspace$ and current action $\at \in \Aspace$. Also, $r: \Sspace \times \Aspace \rightarrow \R$ is the reward function that maps the current state and action to a scalar reward signal, $\gamma \in [0, 1)$ is the discount factor, and $d_0$ defines the initial state distribution $d_0(\state)$.

\subsection{Soft Actor-Critic} 
Soft Actor-Critic~\citep{haarnoja2018soft} follows the policy iteration framework with the objective of maximum-entropy reinforcement learning. It learns a parameterized estimate $Q_\theta(\st, \at)$ of the soft action-value function and a tractable policy $\pi_\phi(\at|\st)$ representing a normal distribution over action space with $\theta$ and $\phi$ being their neural network parameters, respectively. The soft action-value estimate parameters can be obtained by minimizing the following objective:
\begin{align}
J_Q(\theta) = \E{
%(\st, \at, \rtp, \stp)\sim\D}
\D}
{\frac{1}{2}\left(Q_\theta(\st, \at) - \left(R_{t+1} + \gamma 
V_{\hat{\theta}}(\stp)\right)\right)^2}.
\label{eq:q_cost}
\end{align}
The soft-value estimate $V_{\hat{\theta}}(\stp)$ 
can be sampled from
$\E{
\atp\sim\pi_{\phi}
}{Q_{\hat \theta}(\stp, \atp) - \alpha\log\pi_{\phi}(\atp|\stp)}$, where $\alpha$ is the temperature parameter and $\hat \theta$ is the target network parameter. The agent minimizes the following policy objective with reparameterized actions:
\begin{align}
J_\pi(\phi) = \E{\D}{\E{\at\sim\pi_\phi}{\alpha \log\left(\pi_\phi(\at|\st)\right) - Q_\theta(\st, \at)}}.
\label{eq:policy_objective}
\end{align}

\begin{comment}
The temperature $\alpha$ is also automatically adjusted by approximating dual gradient descent:
\begin{align}
J(\alpha) = \E{\mathcal{D}}
{\E{\at\sim \pi_{\phi}}{ - \alpha\log\pi_\phi(\at|\st) - \alpha \hat{\mathcal{H}}}},
\label{eq:ecsac:alpha_objective}
\end{align}
where the target entropy $\hat{\mathcal{H}}$ is a tunable parameter.

\begin{align}
J(\alpha)  = \E{\at\sim \pi_t}{ - \alpha\log\pi_t(\at|\st) - \alpha \hat{\mathcal{H}}},
\label{eq:ecsac:alpha_objective}
\end{align}
where the target entropy $\hat{\mathcal{H}}$ is a tunable parameter.
\end{comment}

\section{TASKS}\label{algorithm}

\begin{figure}[!htbp]
\center
\includegraphics[scale=.85]{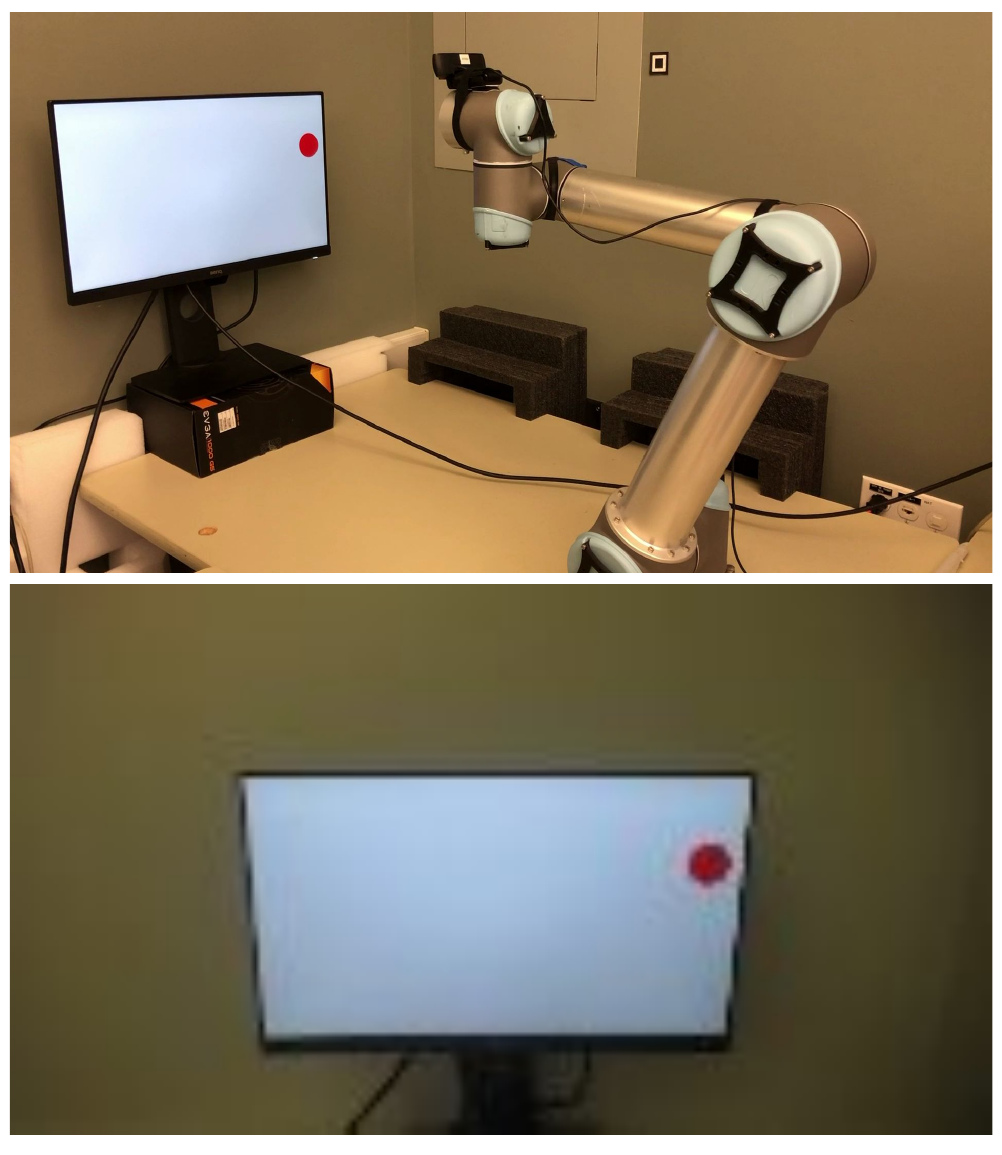}
\caption{
    The setup of the environment (top), and the image from the mounted camera (bottom).
}
\label{fig:environment_setup}
\end{figure}

We develop two vision-based control tasks: \emph{Reaching} and \emph{Tracking}. The objective of the tasks is to reach arbitrary target positions displayed on a computer monitor by the fingertip of the robotic arm using low-level control. Unlike other more complex tasks, such as by~\cite{haarnoja2018soft}, which take about 20 hours for a single run, this simple task setup already fits our research scopes and takes a reasonable amount of time to train. 
The physical setup of the environment is shown in Figure~\ref{fig:environment_setup}  and consists of three devices: 1) a 6-DoF Universal Robotic Arm called UR5, which executes the reaching behavior, 2) an RGB camera, mounted at the end-effector of the arm, which captures images, and 3) a computer monitor, which displays a red target on a white background.
In the real world, time marches on regardless of the time cost of I/O and environment computations so that the information received by the learning agent is always delayed. 
To minimize such systematic delays, we follow the SenseAct framework~\citep{mahmood2018benchmarking}.
These devices communicate with three asynchronous processes, and computations are distributed between the environment process and the three device processes. 

\begin{figure}[!htbp]
\center
\includegraphics[scale=.55, trim=0cm 0 0 0 ]{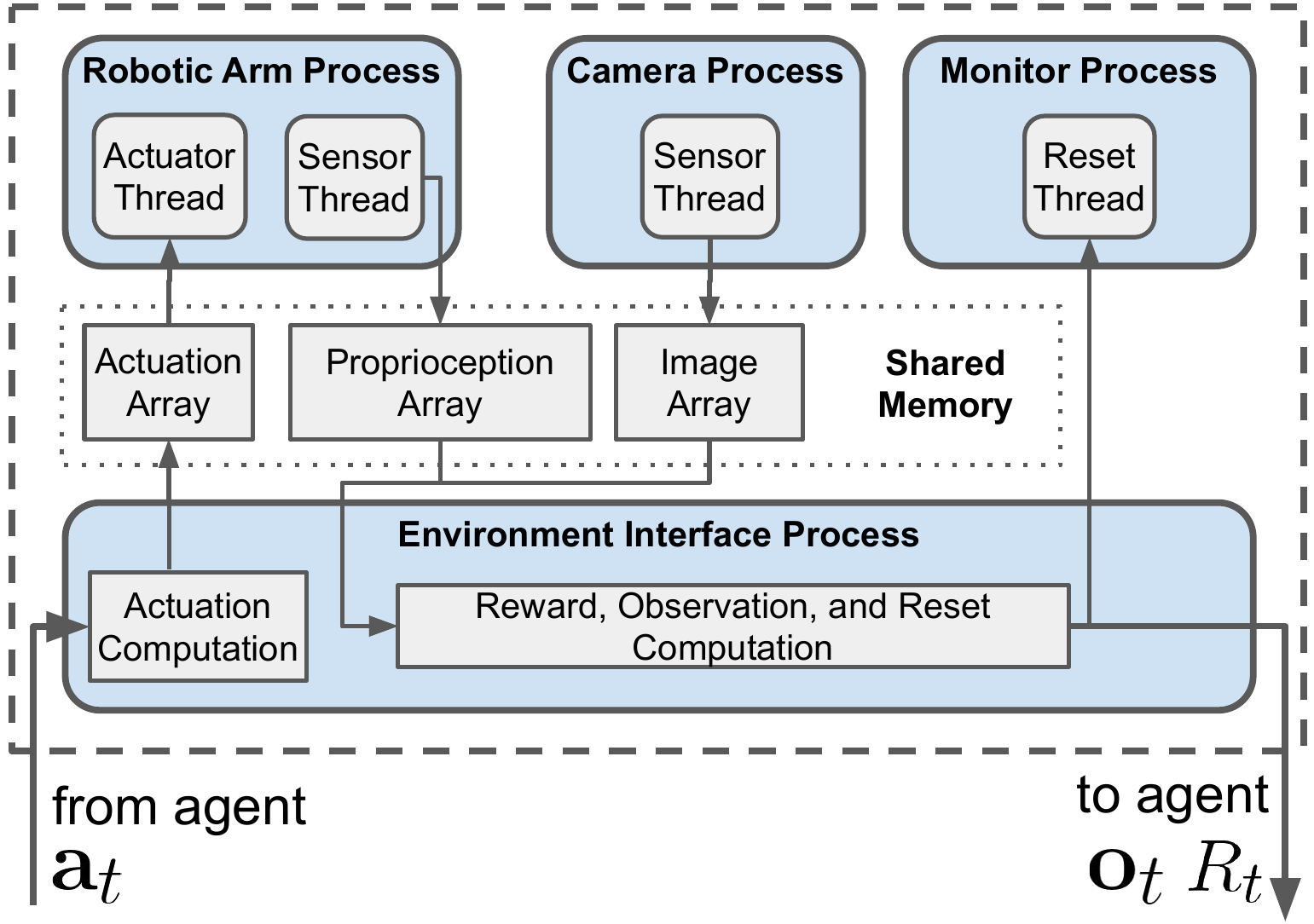}
\caption{
    The environment architecture in this work.
}
\label{fig:env_architecture}
\end{figure}

Figure~\ref{fig:env_architecture}~depicts the architecture of our environment computations. 
The UR5 process sends proprioception of the robotic arm to the environment interface process every 8ms and receives actuation commands from it at its corresponding action cycle time. 
The camera process sends images captured every 40ms. 
The monitor process displays the target and, when applicable, waits for the reset command for randomizing the target location after termination. 
The environment interface process interacts with device processes, computes observations and rewards, and handles the agent-environment interactions. 
All the inter-process communications are through the corresponding I/O threads and shared memory to reduce delayed computations due to data I/O. To synchronize sensory information from different device processes, timestamps are also added to the data stream. 

The observation space includes the most recent three frames of images,  following~\citet{mnih2013playing}, as well as proprioceptive information: current joint angles, current joint velocities, and the previous action, 
where the latter is added to mitigate the negative effect of the non-Markovian property of the environment. 
The image size is either $160 \times 90 \times 3$ or $320 \times 180 \times 3$ in different experimental settings. For the robotic arm, we actuate five joints excluding the wrist, and the action space is the angular velocity between $[-0.7, 0.7]$ rad/s on the five actuated joints. Cartesian boundaries and angle boundaries are both imposed to ensure safety during arm movement. The reward function is defined as
\begin{align}
R_t = \alpha \frac{1}{hw}\sum^{h}_{i=0} \sum^{w}_{j=0} M_{ij} W_{ij}- \beta(|\pi\!-\!\sum_{n=1}^3 \omega_{t,n}|\!+\!|\!\sum_{n=4}^5 \omega_{t,n}|), 
\label{eq:reward_function}
\end{align}
where $h$ and $w$ are the height and width of the image in pixels, $M$ is a 0-1 mask matrix with shape $h \times w$ indicating whether pixels are within the color threshold of the target, and $W$ is the weight matrix with shape $h \times w$ with weights decreasing from 1 to 0 from its center to edges. Finally, $\omega_{t,n}$ is the angle of $n$th joint at step $t$. The first part of the reward function encourages the agent to move closer to the target and keep the target at the center of the frame, while the second part prevents it from twisting the joints too much.
We set the coefficients $\alpha=800$ and $\beta=1$ for all experiments.
The difference between reaching and tracking is whether the targets move and whether they reset.
Specifically,  targets in \textit{Reaching} are randomly generated at the beginning of each episode and stay static during the episode, and targets in \textit{Tracking} move consistently towards a random direction and bounce from the edges.
The episode length is 4s, and the total number of time steps in an episode varies with the chosen action cycle time. After one episode, the environment resets by bringing the arm to a particular position in about 3s.

\section{THE LEARNING ARCHITECTURE}\label{experiment}
Our sequential Soft-Actor-Critic implementation follows~\citet{yarats2019improving}, with two major architectural improvements. First, we use spatial-softmax~\citep{finn2016deep} to convert the encoding vector into soft coordinates to track the target more precisely. Second, we apply random cropping 
%(\citealt{kostrikov2020image}, \citealt{ laskin2020reinforcement}) 
to augment images in mini-batches to learn more robust representations given our limited amount of observations.
\begin{figure*}[!htbp]
\center
\includegraphics[scale=.70]{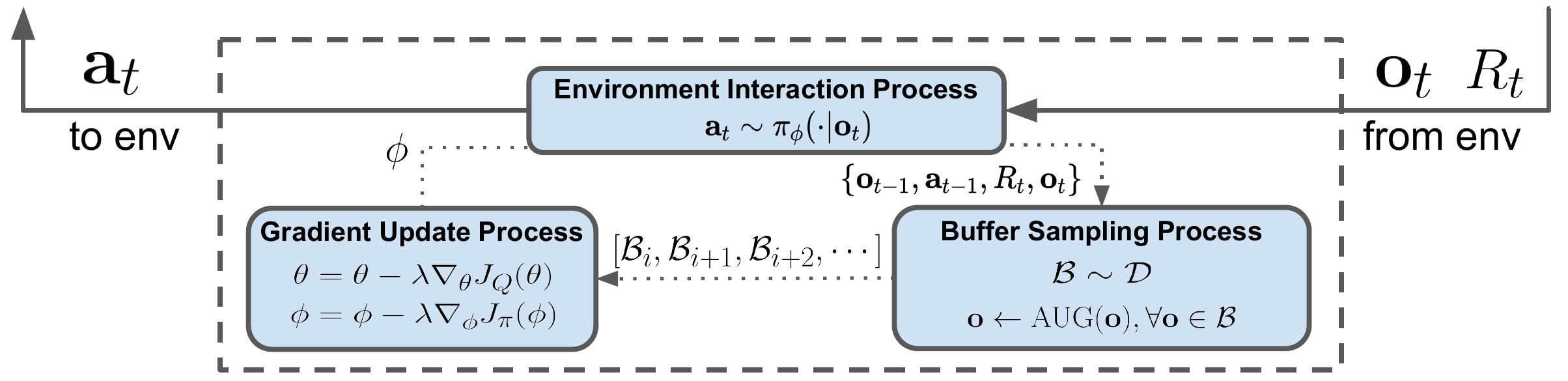}
\caption{
    Overview of Async-SAC-2, our asynchronous learning architecture.
}
\label{fig:architecture}
\end{figure*} 
To enable asynchronous learning updates in Soft-Actor-Critic, we implement an architecture shown in Figure~\ref{fig:architecture}, which can be seen as an extension to an existing architecture~\citep{haarnoja2018learning}. Our architecture consists of three processes that run in parallel: 1) the environment interaction process, which interacts with the environment, collects transition samples, and sends them to the replay buffer, 2) the buffer sampling process, which stores transition data, samples a mini-batch, converts the mini-batch to CUDA tensors and applies random augmentation to it, and 3) the gradient update process, which performs gradient updates and shares parameters with environment interaction process. The existing asynchronous architecture runs learning updates and environment interaction asynchronously, but inside the learning update, the replay buffer sampling and gradient updates are still processed sequentially. 

We extend the existing architecture by separating the replay buffer sampling and gradient updates into two processes and running them in a semi-asynchronous way\footnote{We call it semi-asynchronous as the replay buffer sampling has to happen before the gradient update, but the replay buffer can prepare next samples during the gradient update.}. 
Such semi-asynchronous execution shortens the cycle time for learning updates from the sum of buffer sampling and gradient updates times to the maximal of the two. 
The increase in learning-update frequency is substantial and can be as large as twice when they both take a long and similar amount of time. 
Based on our architecture, we can achieve three different degrees of asynchronous execution. If we run all three components sequentially, it will correspond to most open-source SAC implementations, and we call it \emph{Seq-SAC}. If we only run environment interactions asynchronously but run replay buffer sampling and gradient update sequentially, we will get an architecture similar to~\citet{haarnoja2018learning}, and we call it \emph{Async-SAC-1}. If we run environment interactions asynchronously and run replay buffer sampling and gradient update in parallel, it will correspond to our extended asynchronous learning architecture, and we call it \emph{Async-SAC-2}. 

\begin{figure}[!htbp]
\center
\includegraphics[scale=.45, trim=1.1cm 0 0 0]{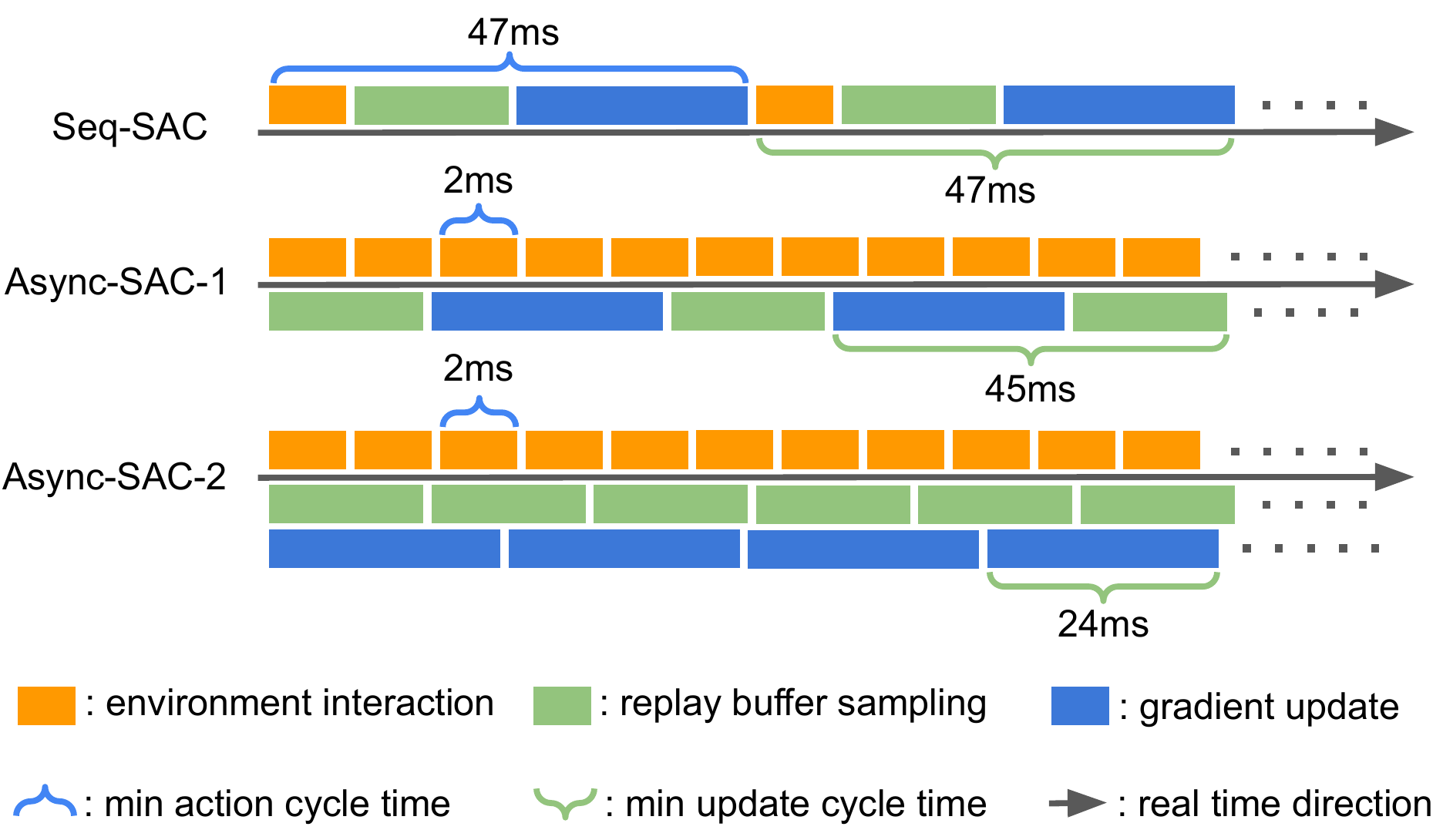}
\caption{\label{fig:computation_flow}
    The computational flow of the three versions of SAC. For drawing purposes, the relative length of each block may not reflect the relative computation time.
}
\end{figure}

The computational flow of the three architectures and the benefit of asynchronous learning in terms of shorter cycle times are shown in Figure~\ref{fig:computation_flow}.
Specifically, we show the computation time spent by environment interaction, replay buffer sampling, and gradient update as the time spent by other operations is negligible in our case.
The amount of times for these computations given around the colored boxes are representative of real computation times, as they are actual measurements in one of our experimental settings. The measurements for all settings are given in Figure~\ref{fig:computation_time}.  From \emph{Seq-SAC} to \emph{Async-SAC-1}, the minimal action cycle time decreases substantially, but the update cycle time almost remains the same. However, from \emph{Async-SAC-1} to \emph{Async-SAC-2}, even the update cycle time decreases significantly. Therefore, our architecture can perform learning updates faster than the existing architecture, which can be crucial when updates are computationally expensive.

\section{THE EXPERIMENTAL SETUP}
\begin{figure}[!htbp]
\center
\includegraphics[scale=.60, trim=0.2cm 0 0 0]{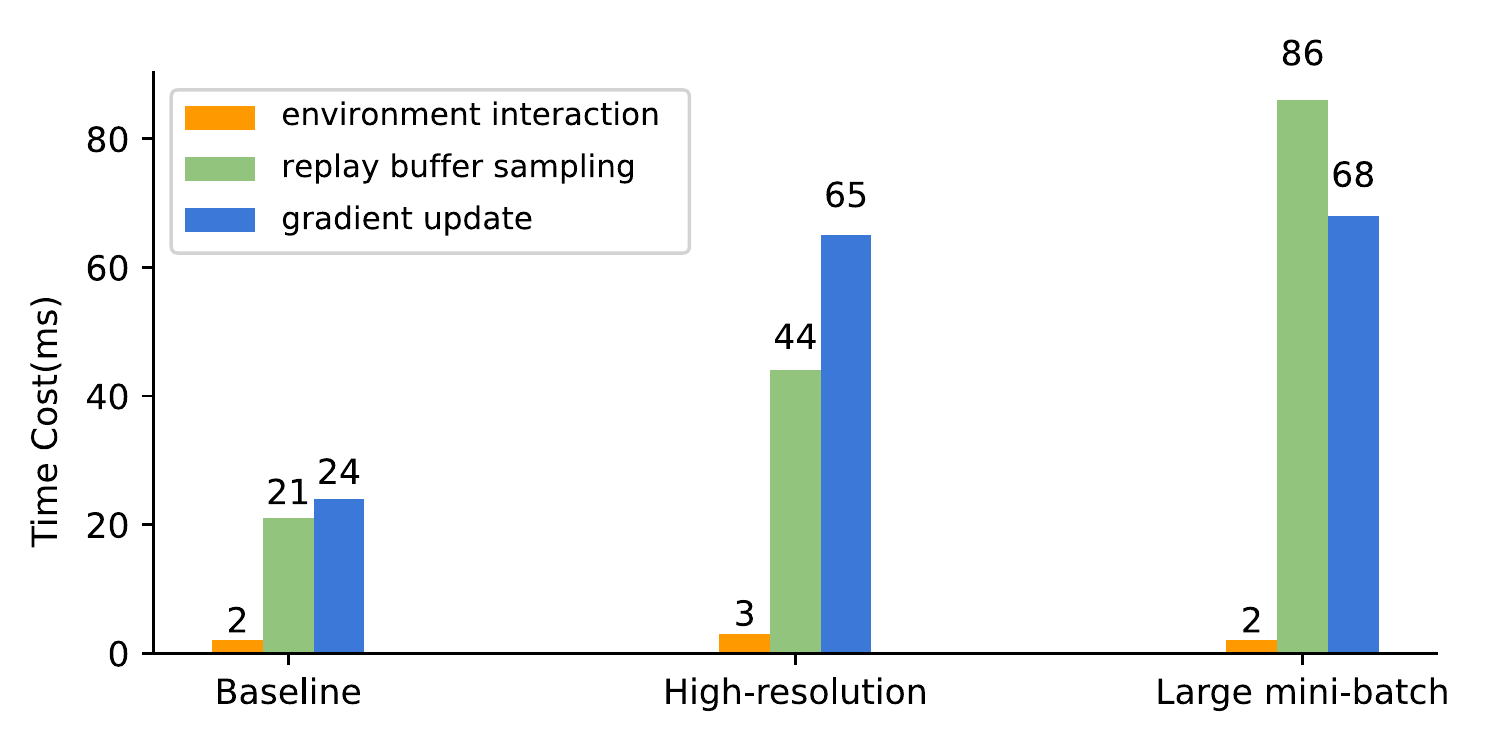}
\caption{
    The computation time of different components under different experiment settings
}
\label{fig:computation_time}
\end{figure}

In our experiments, we compare sequential and asynchronous learning under different settings with different update costs. 
The first setting uses relatively cheap computations, where the mini-batch size is 128 and the image size is 160$\times$90$\times$3. 
The next two settings use increased computational cost. 
In the second settings, the mini-batch size is 128 and image size is 320$\times$180$\times$3. 
In the third settings, the mini-batch size is 512 and image size is 160$\times$90$\times$3. 
We call these three settings: \emph{baseline}, \emph{high-resolution}, and \emph{large mini-batch}, respectively. 
In all settings, both asynchronous learning systems use 40ms action cycle time which is adopted from a prior work on UR5 \citep{mahmood2018setting} while the \emph{Seq-SAC} needs different action cycle times in different settings so that learning updates can fit into them. 
To choose the minimal affordable action cycle time for \emph{Seq-SAC}, we record the time cost of each component in our implementation, as shown in Figure~\ref{fig:computation_time}, and choose the action cycle time accordingly. 
%Our hardware specs are given in Appendix C
As the action cycle time needs to be aligned with the hardware cycle time 8ms of the robotic arm and the cycle time of 40ms of the camera, we use 80ms\footnote{It is possible to use action cycle time 40ms for the sequential SAC, but it will require tiny mini-batch size, which is infeasible for SAC training.} for baseline setting, 120ms for high-resolution setting, and 200ms for large mini-batch setting, considering some extra time needed for the environment interaction.

The episode length of \textit{Reaching} and \textit{Tracking} are both 4 seconds. In this case, the action cycle time of 40ms and 80ms will result in 100 and 50 time steps in one episode, respectively. Because of the variable action cycle time and the constraint of a physical robot, we make the following five modifications to the common experimental methodology. 
First, the variable action cycle time combined with fixed training time steps can lead to substantially different actual training times. To avoid this issue, we use 2-hour wall time as the total training time, regardless of the actual time steps the agent takes. 
Second, the effective time steps in one episode could be shortened due to prolonged action cycle time. Thus, we scale the reward proportional to the action cycle time to make episode returns comparable.
Third, measuring performance by running a separate evaluation phase is incompatible with real-time learning setups. Instead, we report the agent's online performance for evaluation.
Fourth, hyper-parameter search on a physical robot can lead to safety issues and damage the robot. In our experiment, we use the same hyper-parameters across different versions of SAC and different experimental settings.
Fifth, at the beginning of each run, there are 1000 time steps during which the agent executes a random policy to initialize replay buffer. These additional 1000 time steps are neither counted in the 2-hour training time nor shown in the learning curve.

\section{EXPERIMENTAL RESULTS}
\begin{figure*}[!htbp]
\center
\includegraphics[scale=.57
, trim=0.5cm 0 0 0
]{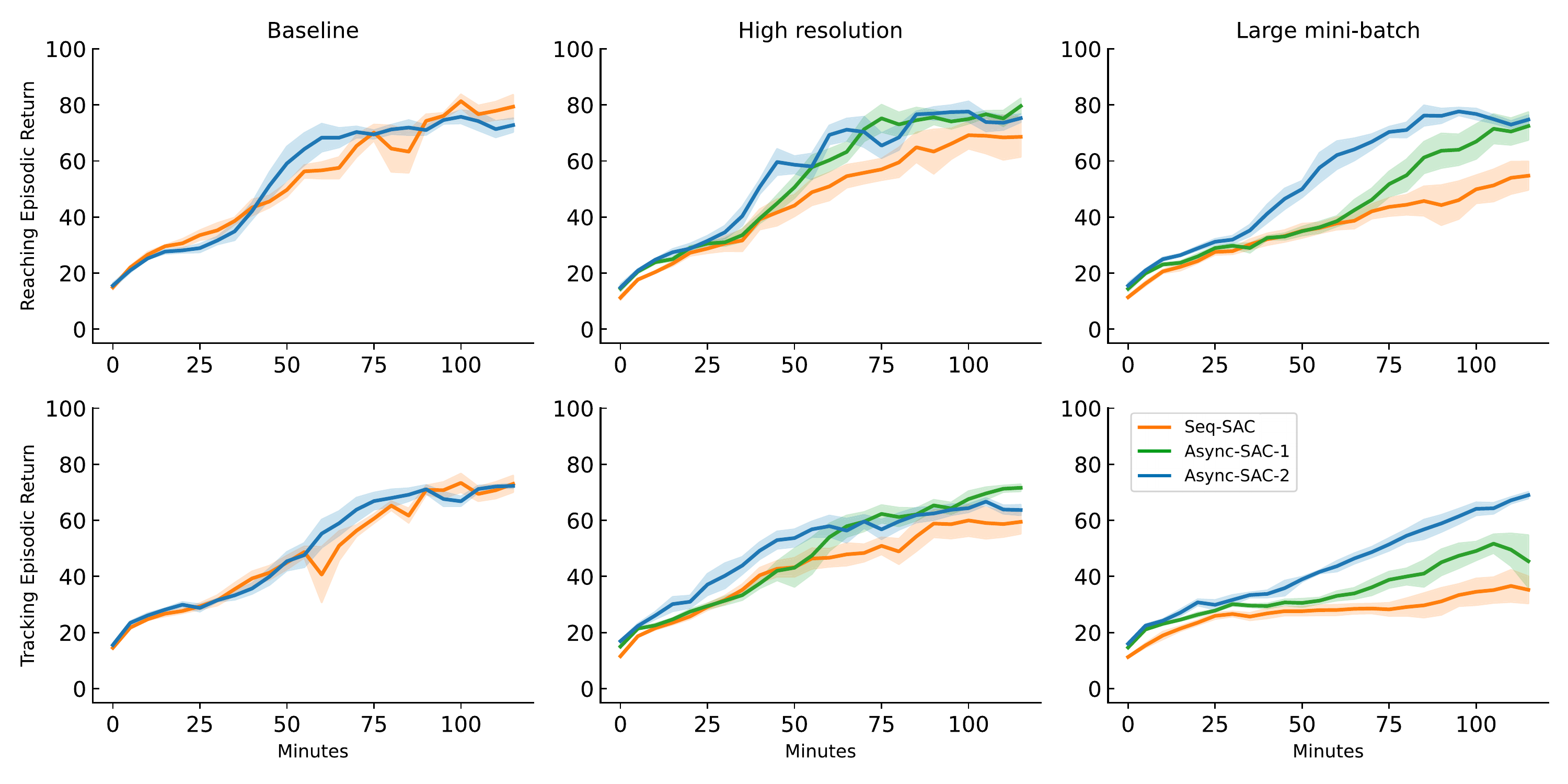}
\caption{
    Learning curves of sequential and asynchronous SACs averaged over five independent runs.
}
\label{fig:experimental_results}
\end{figure*}

\begin{figure}[!htbp]
\center
\includegraphics[scale=.38, trim=1.5cm 0 0 0]{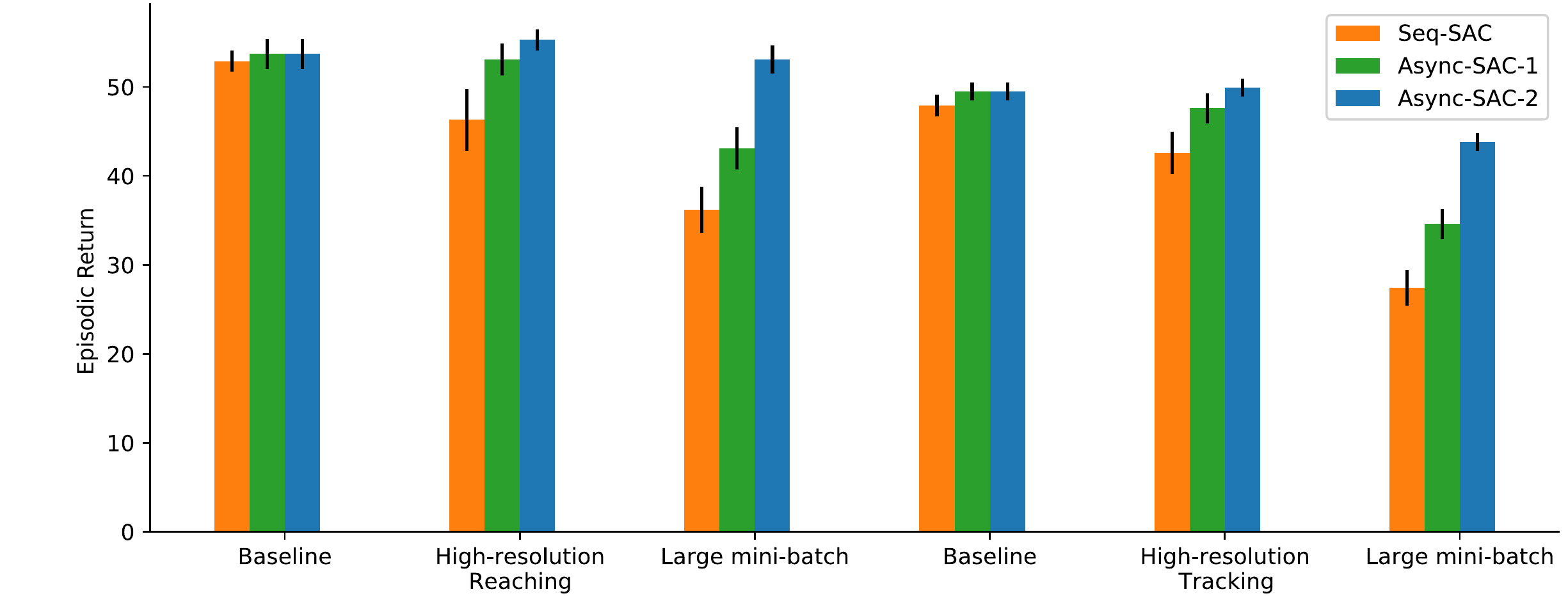}
\caption{
    The overall performance of sequential and asynchronous SAC systems calculated by averaging returns over the whole learning period and five independent runs. 
}
\label{fig:evaluation_results}
\end{figure}

\begin{figure}[!htbp]
\center
\includegraphics[scale=.32]{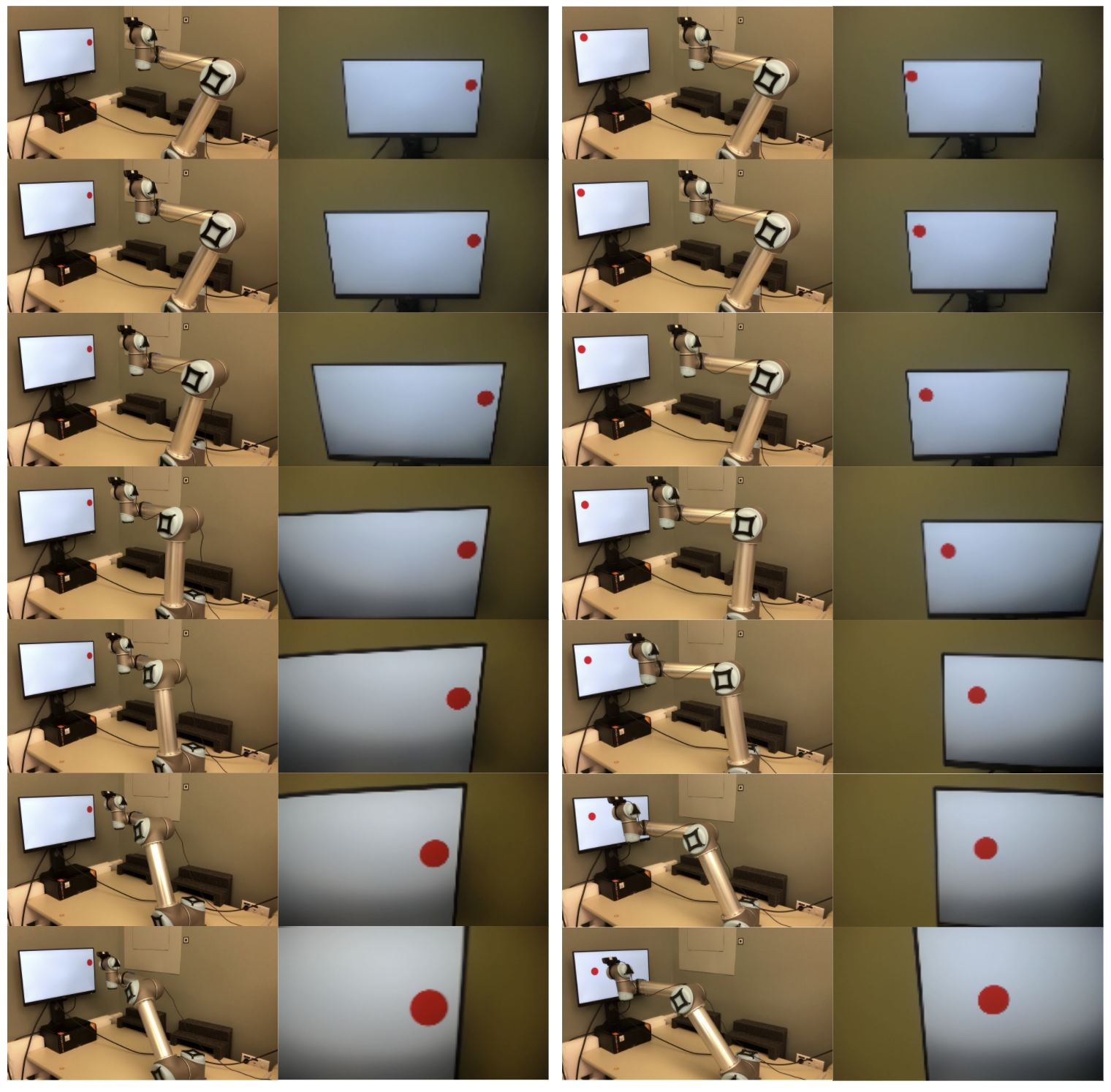}
\caption{
Learned behaviors of \emph{Async-SAC-2} in \textit{Reaching} (left) and \textit{Tracking} (right). A video demo is available at: \url{https://youtu.be/B9icTcV1wNQ}.}
\label{fig:time_lapse}
\end{figure}

We show the learning curves of \emph{Seq-SAC}, \emph{Async-SAC-1}, and \emph{Async-SAC-2} in Figure~\ref{fig:experimental_results}.
In the baseline setting, both versions of asynchronous SAC can make learning updates much faster than the speed of data collection, which could lead to early convergence to sub-optimal policies. 
Thus, we only allowed one learning update per one environment interaction, and \emph{Async-SAC-1} and \emph{Async-SAC-2} were identical in this case. 
Thus we only show one learning curve for the asynchronous SAC in the baseline setting. 
Each learning curve was obtained from 5 independent runs of the corresponding configuration, and the shaded area represents the standard error.
Figure~\ref{fig:evaluation_results} show the overall performance in each of the total six configurations. 
The overall performance is obtained by averaging the returns over the whole period of learning and independent runs for each configuration.
The learned reaching and tracking behaviors by \emph{Async-SAC-2} after 2 hours of training is shown in Figure~\ref{fig:time_lapse}.

Our experiments reveal several interesting properties of real-world reinforcement learning and asynchronous reinforcement learning. In the baseline setting, the main difference between \emph{Seq-SAC} and both versions of asynchronous SAC is the 80ms and 40ms action cycle time. Surprisingly, the sequential learning version performs similarly to the asynchronous learning version. This counter-intuitive result indicates that, though asynchronous learning can enable arbitrarily short action cycle time, which, in principle, could lead to superior policies with finer control, it could also add difficulties to learning due to inconsistent exploration and more bootstrap steps backward.

However, in high-resolution and large mini-batch settings, the action cycle time in \emph{Seq-SAC} has to scale proportionally to the substantially increased computation cost, which leads to excessively long cycle time and starts to impede learning by slowing down the data collection rate and learning update frequency. Meanwhile, from the baseline setting to the large mini-batch setting, the performance degradation is more significant in \textit{Tracking} than in \textit{Reaching}, which indicates that the reduced responsiveness by excessively long action cycle time can make agents perform poorly in non-stationary tasks.

Between the two asynchronous versions, the overall performance of \emph{Async-SAC-2} is only slightly better in high-resolution setting  as shown in Figure~\ref{fig:evaluation_results}. However, in the large mini-batch setting, where the total computation for learning is substantially more than the high-resolution setting, \emph{Async-SAC-2} performed significantly and consistently better than \emph{Async-SAC-1} throughout the learning period in both \textit{Reaching} and \textit{Tracking}.
Overall, our extended architecture is more robust across tasks with different computational costs.

% \newpage
\section{CONCLUSIONS}\label{conclusions}
In this work, we provided the first systematic study on sequential and asynchronous learning for deep reinforcement learning from images with a physical robot. 
In total, we ran 80 independent runs in our experiments, which took nearly 200 hours of usage on the robot. 
In most configurations, effective learning is achieved without tuning the hyper-parameters, which indicates the reliability of the algorithm, our learning architecture, and our task setup. 
We found that the critical benefits of asynchronous learning are more flexibility in choosing action cycle time and better utilization of available computational resources. 
In sequential learning, the action cycle time is bounded by the time cost of learning updates. In contrast, in asynchronous learning, the action cycle time is only constrained by the hardware capabilities. Unfortunately, given such flexibility, it is still unclear how to choose the action cycle time appropriately in a task-specific manner as either too long or too short cycle time could be detrimental to learning. 
Investigating the guidelines of choosing best-performing action cycle time in real-world reinforcement learning remains a promising future research direction. 
When learning sequentially, the computational resource is constantly alternating between idle and busy mode, which causes waste in computation time. 
However, in asynchronous learning, especially based on our architecture, the gradient updates are being processed back to back so that the computation resources are fully utilized. 
However, when gradient update computation is cheap, asynchronous learning can also make gradient updates excessively faster than the environment interactions, which may lead to early convergence or overfitting. 
Devising approaches to avoid this issue while fully utilizing the computation resource is another promising research direction.
\vspace{-0.3cm}
\section*{Acknowledgments}\label{conclusions}
The authors gratefully acknowledge funding from the Huawei Noah’s Ark Lab,
the Canada CIFAR AI Chairs program, 
the Reinforcement
Learning and Artificial Intelligence (RLAI) laboratory,
the Alberta Machine Intelligence Institute (Amii), and
the Natural Sciences and Engineering Research Council (NSERC) of Canada. 
We also thankfully acknowledge the donation of the UR5 from the Ocado Group.

%\addtolength{\textheight}{-12cm}   % This command serves to balance the column lengths
                                  % on the last page of the document manually. It shortens
                                  % the textheight of the last page by a suitable amount.
                                  % This command does not take effect until the next page
                                  % so it should come on the page before the last. Make
                                  % sure that you do not shorten the textheight too much.

% References are important to the reader; therefore, each citation must be complete and correct. If at all possible, references should be commonly available publications.

%\newpage
\bibliographystyle{apacite} %% setting the cite style
\bibliography{bibs}

\end{document}